\documentclass[11pt,a4paper]{article}

\usepackage[margin=1in]{geometry}
\usepackage{graphicx}
\usepackage{multirow}
\usepackage{amsmath,amssymb,amsfonts}
\usepackage{booktabs}
\usepackage{url}
\usepackage[round]{natbib}
\usepackage{hyperref}
\usepackage{setspace}
\begin{document}

\title{Instruction Complexity Induces Positional Collapse in Adversarial LLM Evaluation}

\author{Jon-Paul Cacioli\\
ORCID: 0009-0000-7054-2014\\
Independent Researcher, Melbourne, Australia\\
\texttt{synthium@hotmail.com}}

\date{}

\maketitle

\begin{abstract}
When instructed to underperform on multiple-choice evaluations, do language models engage with question content or fall back on positional shortcuts? We map the boundary between these regimes using a six-condition adversarial instruction-specificity gradient administered to two instruction-tuned LLMs (Llama-3-8B and Llama-3.1-8B) on 2,000 MMLU-Pro items. Distributional screening (response-position entropy) and an independent content-engagement criterion (difficulty-accuracy correlation) jointly characterise each condition. The gradient reveals three regimes rather than a monotonic transition. Vague adversarial instructions produce moderate accuracy reduction with preserved content engagement. Standard sandbagging and capability-imitation instructions produce positional entropy collapse with partial content engagement. A two-step answer-aware avoidance instruction produces extreme positional collapse, with near-total concentration on a single response position (99.9\% and 87.4\%) and no measurable content sensitivity. This was the only multi-step instruction tested, and it produced the most extreme shortcut. The attractor position matches each model's content-absent null-prompt default. The effect replicates across both models and four academic domains. Distributional collapse and content engagement can co-occur (50\% concordance between screening criteria), indicating that entropy-based screening and difficulty-based content assessment capture partially independent dimensions of response validity. Results suggest that instruction complexity can determine whether adversarial compliance uses content-aware or content-blind mechanisms in small instruction-tuned LLMs under greedy decoding.
\end{abstract}

\noindent\textbf{Keywords:} LLM evaluation, sandbagging detection, positional bias, psychometrics, validity screening, instruction following

\section{Introduction}\label{sec:intro}

\subsection{Background}\label{sec:background}

Sandbagging is the deliberate suppression of capability on evaluations. It is a recognised threat to AI safety assessment \citep{vanderweij2025,meinke2024}. Detection methods range from white-box linear probes \citep{nguyen2025} to black-box accuracy comparisons \citep{taylor2025}, each with known limitations. A complementary approach treats sandbagging detection as a validity screening problem. Before interpreting what a model's evaluation responses mean, first check whether the response pattern is valid. That is, whether responses reflect item-level engagement rather than a content-blind structural artefact.

This validity-first approach has deep roots in clinical psychometrics. The MMPI-3 screens for response validity (L, F, K scales) before interpreting clinical scales \citep{benporath2020}. The PAI enforces a similar two-stage architecture \citep{morey2007}. The logic is domain-independent. If a response pattern is dominated by a response set that overrides item content, downstream analyses are fitting noise. The same statistical problem arises in LLM evaluation when models satisfy adversarial instructions via positional shortcuts rather than content-aware strategies. LLMs exhibit substantial positional biases in multiple-choice settings even under honest conditions \citep{pezeshkpour2024,zheng2024}. The present study extends this by showing that positional bias is not merely a baseline nuisance but an active mechanism through which models satisfy adversarial instructions. In psychometric terms, response-position entropy functions as a response-style indicator. The difficulty-accuracy correlation functions as a measure of item discrimination. Positional collapse corresponds to non-content-based responding.

Two prior studies established the empirical basis. Study~1 \citep{cacioli2026a} discovered that Llama-3-8B exhibited positional collapse under explicit sandbagging instruction. Response-position entropy dropped from 0.977 to 0.793, with 58\% of responses concentrated on options E and F. A dissociation was observed. A specific ``least likely'' instruction produced content-aware behaviour. A vague ``deliberately underperform'' instruction produced the positional shortcut. Study~2 \citep{cacioli2026b} confirmed via cyclic option-order permutation that this collapse is a model-level distributional attractor, not a dataset artefact (same-letter rate $r = 0.9994$, JS divergence $= 0.027$ between original and permuted orderings).

These findings raise a precise question. Where, along the continuum from vague to specific adversarial instruction, does the model transition from content-blind shortcut to content-aware strategy? Does this transition follow a monotonic gradient, or does instruction complexity introduce non-monotonic effects?

\subsection{Related work}\label{sec:related}

The shortcut learning literature provides the conceptual framework. \citet{geirhos2020} established that neural networks systematically prefer simpler decision rules when available, even when more complex content-aware strategies would better satisfy the objective. Applied to adversarial instruction-following, this predicts that models will default to the simplest compliant strategy unless the instruction makes content engagement unavoidable.

Recent work on instruction-following clarifies why instruction complexity matters. \citet{lim2025} demonstrated an ``atomic instruction gap.'' Instruction-tuned models struggle with simple, self-contained directives that require precise execution. \citet{yang2025} showed that underspecified prompts are fragile and approximately twice as likely to produce regressive behaviour. \citet{baxi2025} identified orthogonal failure modes, constraint compliance versus semantic accuracy, that parallel the content-blind versus content-aware distinction in the present study.

In the sandbagging literature, \citet{li2025} taxonomised covert sandbagging policies (false-explanation, answer-swap). \citet{taylor2025} demonstrated that accuracy-based black-box detection fails against effective capability imitation. The present study targets a different detection channel. Distributional structure of response positions is orthogonal to accuracy level and detectable even when accuracy imitation succeeds.

\subsection{The present study}\label{sec:present}

We tested seven hypotheses in a pre-registered study (OSF: \url{https://osf.io/7p64}).

\textbf{H1} (gradient). Instruction specificity across six adversarial conditions (S1 to S6) systematically shifts responding from content-blind shortcut regime to content-aware regime. The transition may be non-monotonic at intermediate levels.

\textbf{H2} (shortcut regime). Vague (S1) and standard sandbagging (S2) instructions produce entropy collapse, high modal concentration, and near-zero difficulty-accuracy correlation.

\textbf{H3} (content-aware regime). Answer-aware avoidance (S5) and least-likely selection (S6) produce near-uniform distributions with significant difficulty-accuracy correlation.

\textbf{H4} (transition zone). Capability imitation (S3) and direct inversion (S4) produce intermediate patterns.

\textbf{H5} (cross-model). Llama-3.1-8B shows earlier transition to content-aware regime than Llama-3-8B.

\textbf{H6} (screening concordance). Distributional indicators and difficulty-accuracy correlation concordantly classify conditions.

\textbf{H7} (S5 reversion). S5 may produce partial reversion to shortcut behaviour due to two-step reasoning demands.

\section{Method}\label{sec:method}

The study was pre-registered on OSF (\url{https://osf.io/7p64}) prior to data collection. No deviations from the pre-registration occurred. All code and data are available at \url{https://github.com/synthiumjp/bcb-sandbagging-pilot}.

\subsection{Design}\label{sec:design}

A fully crossed within-item repeated measures design. Two models received seven primary conditions and four null-prompt baselines on the same items, yielding 29,600 trials. Each trial used an independent conversation context, eliminating carryover effects between items.

\subsection{Models}\label{sec:models}

Llama-3-8B-Instruct and Llama-3.1-8B-Instruct (Meta), both at Q5\_K\_M quantisation. All inference via llama-cpp-python 0.3.16 with Vulkan backend on an AMD Radeon RX 7900 GRE (16~GB VRAM). Greedy decoding ($T=0$, $K=1$) with independent conversation context per item.

\subsection{Dataset}\label{sec:dataset}

2,000 items from MMLU-Pro \citep{wang2024}, stratified by domain (500 each from physics, law, psychology, economics), sampled with seed 42. All items had exactly 10 options (A to J). This is the identical item set used in Studies~1 and~2.

\subsection{Conditions}\label{sec:conditions}

Seven primary conditions, ordered by instruction specificity. Table~\ref{tab:conditions} characterises each condition's operational structure.

\begin{table}[ht]
\centering
\caption{Condition characteristics.}\label{tab:conditions}
\begin{tabular}{llccc}
\toprule
Cond. & Label & Steps & Latent correct ans. & Self-ref. \\
\midrule
H  & Honest baseline        & 1 & No        & No  \\
S1 & Vague adversarial      & 1 & No        & No  \\
S2 & Sandbagging            & 1 & No        & No  \\
S3 & Capability imitation   & 1 & Partially & Yes \\
S4 & Direct inversion       & 1 & Yes       & No  \\
S5 & Answer-aware avoidance & 2 & Yes (maintained) & Yes \\
S6 & Least likely           & 1 & Yes       & No  \\
\bottomrule
\end{tabular}
\end{table}

S5 is the only condition requiring two sequential operations on the same item. It requires identifying the most likely correct answer, then selecting against it. This requires maintaining an intermediate representation across steps. No other condition has this structural demand. S6 also requires a latent correct-answer representation (relative likelihood estimation over options), but it is a single-pass operation. The model does not need to hold one answer in memory while selecting a different one. This distinction between single-pass evaluation and sequential maintain-then-avoid is the key structural difference between S5 and S6.

Four null-prompt baselines (200-item subset, seed 43). N1: random letter, no question. N2: pick any option. N3: format-only, no question content. N4: honest prompt with nonsense question stem and real options.

\subsection{Measures}\label{sec:measures}

\textbf{Primary screening index.} Normalised response-position entropy, computed as $H(\text{responses}) / \log(10)$. Values near 1.0 indicate uniform responding. Values near 0 indicate positional concentration.

\textbf{Independent content-engagement criterion.} Spearman $\rho$ between item difficulty ($1 - $ proportion correct under H) and item-level correctness under each adversarial condition. Near-zero $\rho$ indicates content-blind responding. Significant negative $\rho$ indicates content-engaged responding.

\textbf{Distributional distance.} Jensen-Shannon divergence and total variation distance from honest baseline.

\textbf{Concentration measures.} Modal concentration (proportion at modal position) and top-3 concentration.

\textbf{Accuracy by correct-answer position.} Accuracy stratified by which position (A to J) holds the correct answer, with OLS slope across positions.

\subsection{Analysis plan}\label{sec:analysis}

Primary: Spearman trend test across S1 to S6 for entropy and difficulty-accuracy $\rho$. Pairwise bootstrap 95\% CIs for adjacent condition contrasts (10,000 resamples). Secondary: cross-tabulation of distributional screening verdict against difficulty-accuracy criterion (descriptive). Tertiary: mixed-effects logistic regression, correctness predicted by position, condition, and their interaction, with random intercept for item. All tests at $\alpha = 0.01$.

\section{Results}\label{sec:results}

Parse failure rates were negligible across all conditions (maximum 3/2000 = 0.15\%), well below the pre-registered 5\% threshold.

\subsection{The instruction-specificity gradient}\label{sec:gradient}

\begin{figure}[ht]
\centering
\includegraphics[width=\textwidth]{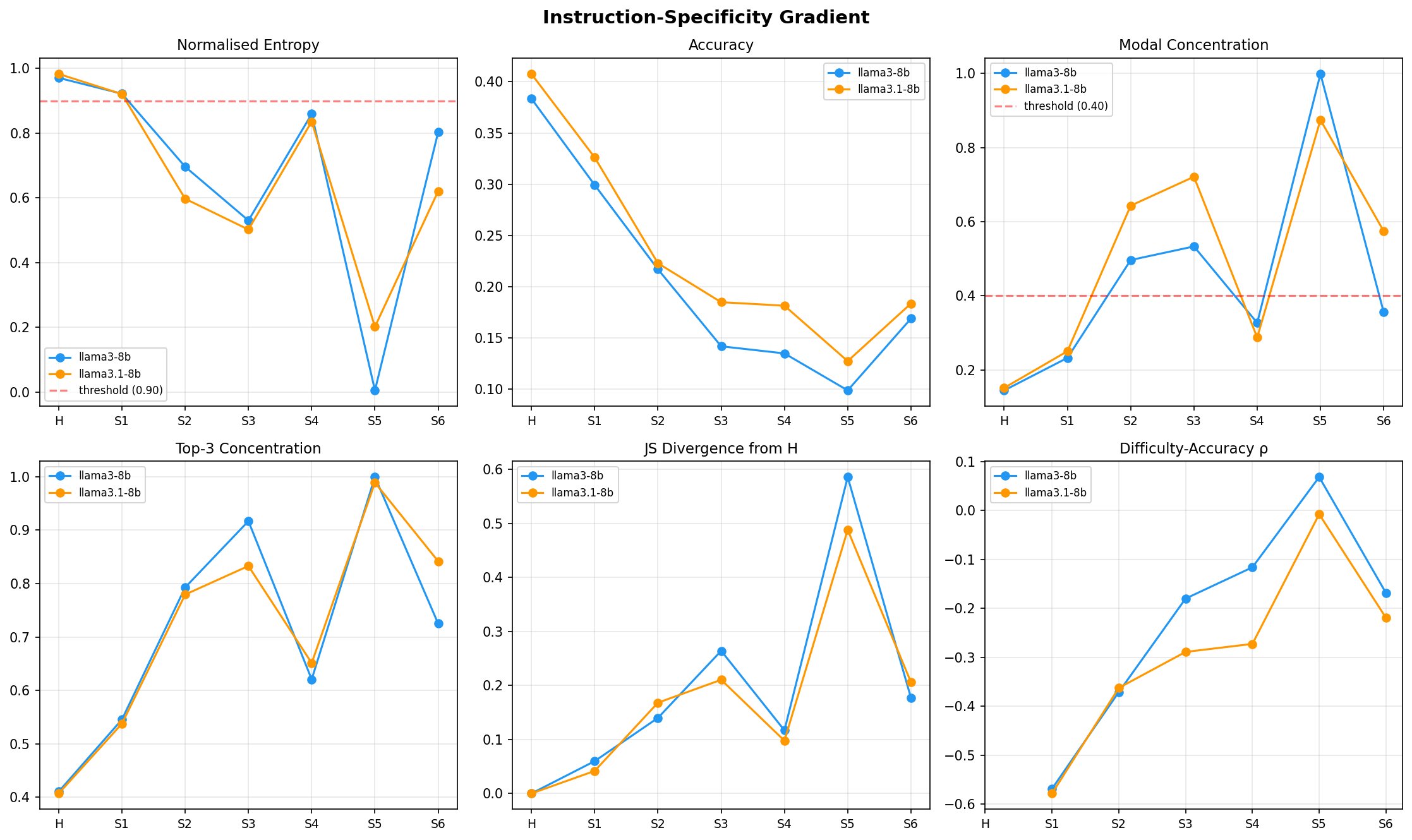}
\caption{Instruction-specificity gradient across six measures for both models. The S5 collapse is visible in every panel. Dashed lines indicate pre-registered thresholds (entropy 0.90, modal concentration 0.40).}\label{fig:gradient}
\end{figure}

Table~\ref{tab:measures} presents all pre-registered measures. Figure~\ref{fig:gradient} shows the gradient visually. The gradient is non-monotonic and consistent across both models.

\begin{table}[ht]
\centering
\caption{Summary measures for all primary condition-model cells.}\label{tab:measures}
\small
\begin{tabular}{llrrrrrrr}
\toprule
Model & Cond & Acc & Entropy & Modal\% & JS & TV & $\rho$ & Mode \\
\midrule
3-8B  & H  & .384 & .971 & .144 & .000 & .000 & ---    & B \\
3-8B  & S1 & .299 & .922 & .232 & .060 & .309 & $-.569$ & G \\
3-8B  & S2 & .217 & .696 & .496 & .140 & .466 & $-.371$ & E \\
3-8B  & S3 & .142 & .530 & .533 & .264 & .626 & $-.180$ & E \\
3-8B  & S4 & .135 & .859 & .327 & .117 & .411 & $-.116$ & G \\
3-8B  & S5 & .099 & .006 & .999 & .586 & .947 & $+.069$ & I \\
3-8B  & S6 & .169 & .803 & .355 & .177 & .524 & $-.168$ & I \\
\midrule
3.1-8B & H  & .408 & .983 & .151 & .000 & .000 & ---    & C \\
3.1-8B & S1 & .326 & .921 & .250 & .042 & .236 & $-.579$ & G \\
3.1-8B & S2 & .223 & .597 & .643 & .168 & .519 & $-.363$ & G \\
3.1-8B & S3 & .185 & .502 & .721 & .211 & .597 & $-.289$ & G \\
3.1-8B & S4 & .182 & .836 & .287 & .098 & .390 & $-.273$ & G \\
3.1-8B & S5 & .128 & .202 & .874 & .488 & .844 & $-.007$ & I \\
3.1-8B & S6 & .184 & .621 & .574 & .206 & .545 & $-.219$ & J \\
\bottomrule
\end{tabular}
\end{table}

\textbf{S1 (vague adversarial).} Minimal distributional disruption. Entropy remained high (0.922/0.921). Accuracy decreased modestly (29.9\%/32.6\% vs.\ 38.4\%/40.8\% honest). Difficulty-accuracy $\rho$ was strongly negative ($-0.569$/$-0.579$), indicating preserved content engagement.

\textbf{S2 (sandbagging).} Positional collapse matching Studies~1 and~2. Entropy dropped to 0.696/0.597. Modal concentration reached 49.6\% at E (Llama-3-8B) and 64.3\% at G (Llama-3.1-8B). Difficulty-accuracy $\rho$ remained significantly negative ($-0.371$/$-0.363$, $p < 10^{-60}$). Despite using a positional shortcut, the model retained difficulty sensitivity.

\textbf{S3 (capability imitation).} Deeper collapse (entropy 0.530/0.502, modal concentration 53.3\%/72.1\%). Both models achieved accuracy within the pre-registered 5--20\% tolerance band (14.2\%/18.5\%). Difficulty-accuracy $\rho$ weakened but remained significant ($-0.180$/$-0.289$).

\textbf{S4 (direct inversion).} Entropy partially recovered (0.859/0.836), with broader response distributions. Difficulty-accuracy $\rho$ was $-0.116$/$-0.273$, still significantly negative.

\textbf{S5 (answer-aware avoidance).} The most extreme result in the study. Entropy collapsed to 0.006 for Llama-3-8B (99.9\% of responses at position I) and 0.202 for Llama-3.1-8B (87.4\% at I). Difficulty-accuracy $\rho$ was $+0.069$ for Llama-3-8B (bootstrap 95\% CI [$+0.027$, $+0.109$]) and $-0.007$ for Llama-3.1-8B (CI [$-0.051$, $+0.037$]). Both are effectively zero.

\begin{figure}[ht]
\centering
\includegraphics[width=\textwidth]{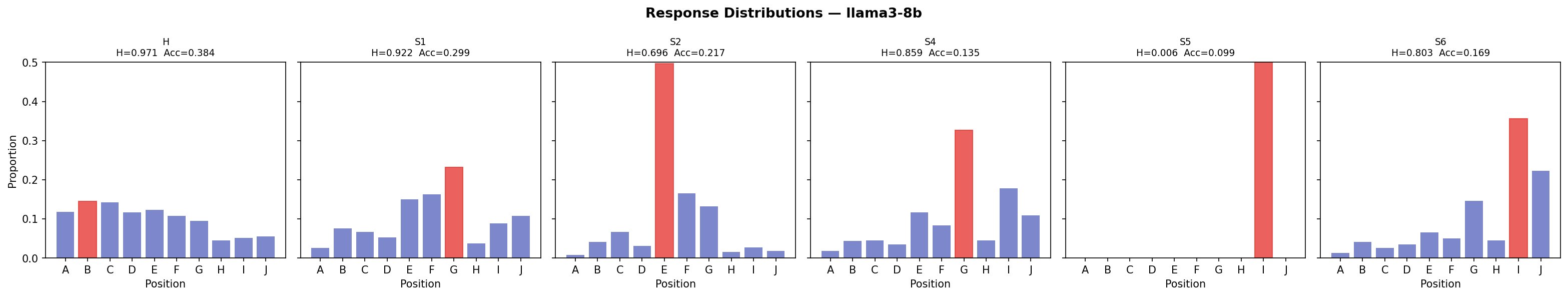}\\[6pt]
\includegraphics[width=\textwidth]{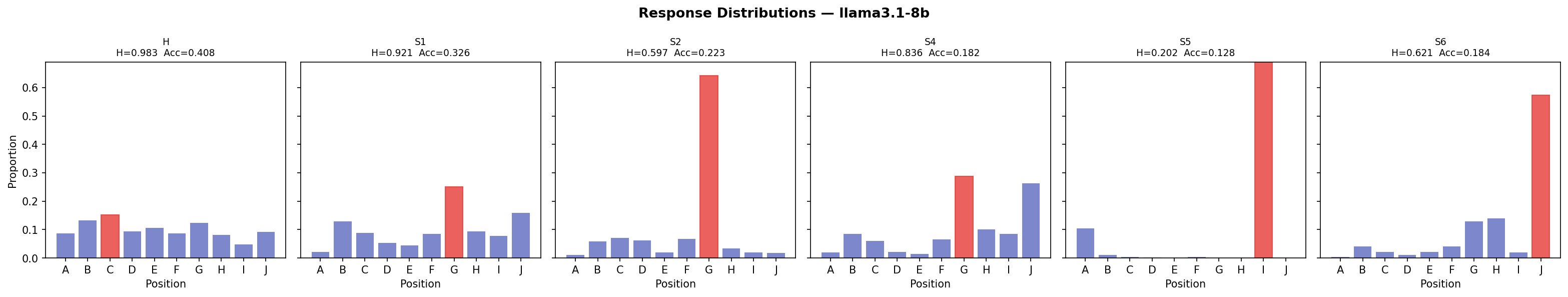}
\caption{Response-position distributions for selected conditions. Top: Llama-3-8B. Bottom: Llama-3.1-8B. Red bars indicate modal position. S5 shows near-total concentration at position I for both models.}\label{fig:distributions}
\end{figure}

\begin{figure}[ht]
\centering
\includegraphics[width=\textwidth]{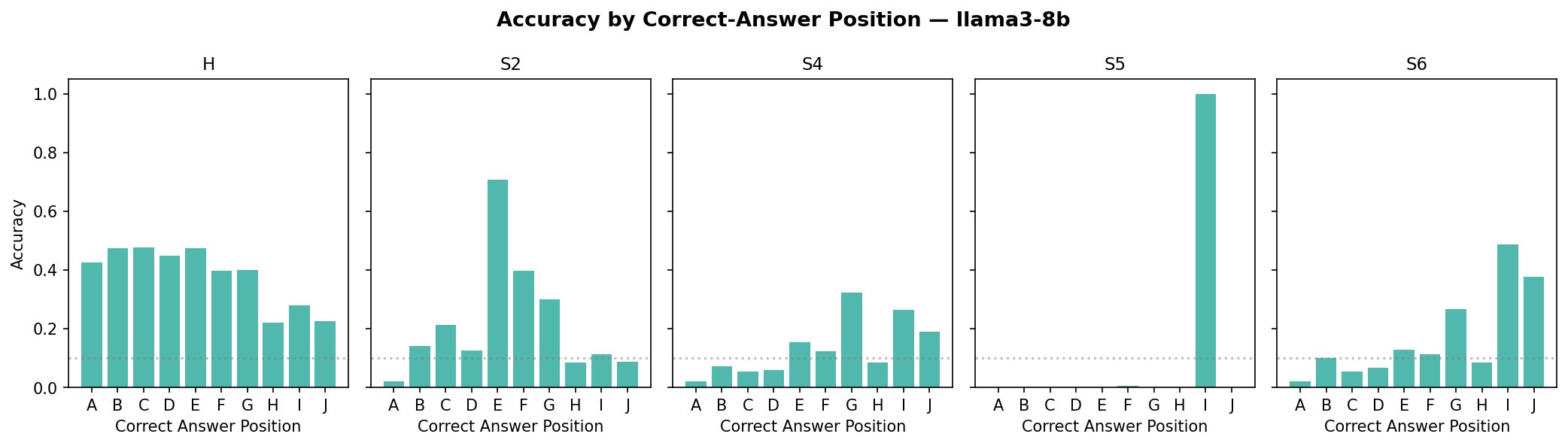}\\[6pt]
\includegraphics[width=\textwidth]{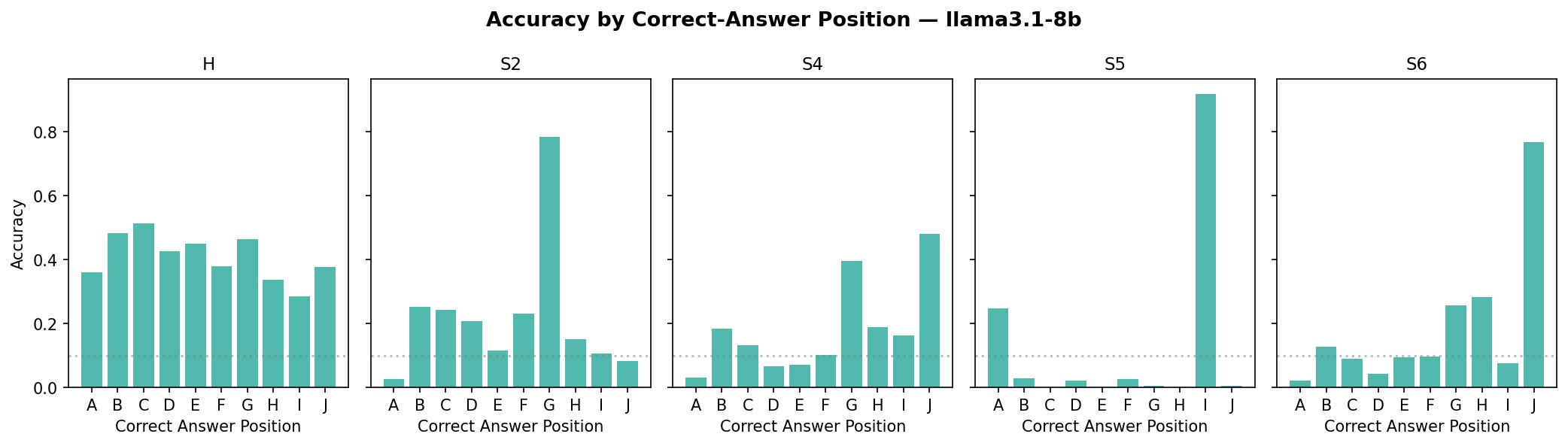}
\caption{Accuracy by correct-answer position. Top: Llama-3-8B. Bottom: Llama-3.1-8B. Under S5, accuracy spikes only when the correct answer occupies the attractor position (I). Dashed line indicates chance (0.10).}\label{fig:accbypos}
\end{figure}

\textbf{S6 (least likely).} Partial recovery from the S5 collapse. Entropy was 0.803/0.621. Difficulty-accuracy $\rho$ was significantly negative ($-0.168$/$-0.219$). Figure~\ref{fig:distributions} shows the response distributions for selected conditions. Figure~\ref{fig:accbypos} shows accuracy stratified by correct-answer position.

\subsection{Trend analysis}\label{sec:trend}

The Spearman trend for entropy across S1 to S6 was non-significant ($\rho = -0.371$, $p = 0.47$ for both models). The gradient is not monotonic. Bootstrap pairwise contrasts confirmed that every adjacent transition was significant (all CIs excluding zero). The S4 to S5 transition was the largest (Llama-3-8B: $-0.854$, CI [$-0.868$, $-0.837$]; Llama-3.1-8B: $-0.634$, CI [$-0.657$, $-0.610$]).

The Spearman trend for difficulty-accuracy $\rho$ was positive and significant for Llama-3.1-8B ($\rho = 0.943$, $p = 0.005$) but not for Llama-3-8B ($\rho = 0.829$, $p = 0.042$ at $\alpha = 0.01$).

\subsection{H7: S5 collapse}\label{sec:h7}

H7 predicted partial reversion at S5. The observed effect far exceeded partial. The S4 to S5 entropy drop was $-0.854$/$-0.634$, producing the most extreme positional collapse in the dataset.

The S5 attractor position (I) matched the content-absent null baseline N3 (format-only prompt) for both models. N3 also produced 100\% concentration at I. Under S5, the model's response distribution is indistinguishable from its behaviour when no question content is presented.

This observation is consistent with multiple mechanisms. Instruction-following failure, where the model cannot decompose the two-step operation. Representation failure, where the model cannot maintain the correct-answer representation across steps. Token-level positional priors, where the instruction maps to a default token preference. Decoding artefacts specific to Q5\_K\_M quantisation. The present design cannot distinguish between these. What can be concluded is that S5 produced content-blind responding indistinguishable from the content-absent baseline. No single-step condition produced comparable collapse.

\subsection{Screening concordance}\label{sec:concordance}

The pre-registered concordance analysis cross-tabulated the distributional screening verdict (entropy $< 0.90$ AND modal concentration $> 0.40$) against the independent difficulty-accuracy criterion ($|\rho| < 0.05$). Agreement was 50\% for both models (3/6 conditions concordant). The discordance is itself the primary finding. It reveals that distributional screening and content engagement measure partially independent dimensions of response validity. Models frequently adopt positional shortcuts (entropy collapse) without fully abandoning semantic processing (difficulty sensitivity).

The discordant cases consistently showed distributional collapse co-occurring with significant content engagement. Only S5 for Llama-3.1-8B showed both distributional collapse and content-blindness ($\rho = -0.007$, $|\rho| < 0.05$).

Sensitivity analysis across threshold bands (entropy 0.85--0.95, modal 0.35--0.50, $|\rho|$ 0.03--0.07) confirmed that discordance is a stable feature of the data. Concordance ranged from 33\% to 83\%.

\subsection{Null baselines}\label{sec:null}

N1 (random letter) and N3 (format-only) produced 100\% concentration at a single position (E and I respectively) across both models. N2 (pick any) and N4 (nonsense stem with real options) produced near-uniform distributions (entropy 0.907--0.960). The false-positive rate for the entropy screen under content-present null prompts (N2, N4) was 0/4.

\subsection{Sensitivity analyses}\label{sec:sensitivity}

Difficulty-accuracy correlations recomputed using pooled difficulty (averaged across both models) differed from within-model estimates by a maximum of 0.031. S3 accuracy was within the pre-registered 5--20\% band for both models (14.2\%/18.5\%). Under honest conditions, the accuracy-by-position slope was negative ($-0.028$/$-0.013$). Under S5, slope reversed to positive ($+0.043$/$+0.024$).

\subsection{Position by condition interaction}\label{sec:mixed}

The pre-registered mixed-effects logistic regression confirmed that every adversarial condition significantly altered the position-accuracy relationship (Table~\ref{tab:mixed}). Under honest conditions, position had a negative main effect on accuracy (Llama-3-8B: $\beta = -0.027$, $p < 0.001$; Llama-3.1-8B: $\beta = -0.012$, $p < 0.001$). All six position $\times$ condition interactions were positive and significant (all $p < 0.001$). Under S5, the net position effect was positive (Llama-3-8B: $-0.027 + 0.069 = +0.042$).

\begin{table}[ht]
\centering
\caption{Position $\times$ condition interaction coefficients (mixed-effects logistic regression).}\label{tab:mixed}
\begin{tabular}{lcc}
\toprule
Interaction & 3-8B $\beta$ & 3.1-8B $\beta$ \\
\midrule
position $\times$ S1 & $+0.038$ & $+0.027$ \\
position $\times$ S2 & $+0.029$ & $+0.020$ \\
position $\times$ S3 & $+0.040$ & $+0.025$ \\
position $\times$ S4 & $+0.051$ & $+0.043$ \\
position $\times$ S5 & $+0.069$ & $+0.034$ \\
position $\times$ S6 & $+0.067$ & $+0.059$ \\
\bottomrule
\end{tabular}\\[4pt]
\footnotesize All interactions $p < 0.001$. Reference: H. Random intercept for item. $N = 13{,}996$ / $13{,}992$.
\end{table}

\section{Discussion}\label{sec:discussion}

\subsection{Why does S5 collapse?}\label{sec:why}

S5 produced more extreme positional collapse than any other condition. Several candidate mechanisms could produce this pattern.

First, the two-step operation may exceed the model's capacity for instruction decomposition at the 8B scale under greedy decoding. Instruction-tuned models at this scale struggle with multi-step directives even in non-adversarial settings \citep{lim2025}. S5 is the only condition requiring sequential operations on the same item (Table~\ref{tab:conditions}).

Second, the adversarial self-referential structure of S5 may create representational interference. The instruction simultaneously activates truthful answering (to identify the correct answer) and adversarial compliance (to avoid that answer). At the 8B scale, these competing activations may not be resolvable, producing collapse to a default prior.

Third, the collapse may reflect a token-level or decoding-level artefact. The attractor position (I) could reflect a token-ID prior amplified by Q5\_K\_M quantisation. However, the same position emerges under N3 (content-absent null). This suggests it is the model's general default rather than a condition-specific artefact.

The present design cannot distinguish between these mechanisms. What it establishes is the empirical regularity. The only multi-step instruction coincides with the most extreme content-blind collapse. This collapse matches the model's content-absent baseline. Only one multi-step instruction was tested, so the claim that multi-step structure per se drives collapse remains a hypothesis for future work. Chain-of-thought prompting, instruction decomposition ablations, rephrased S5 variants, or activation-level probes could disambiguate the causal mechanism.

\subsection{The entropy-rho dissociation}\label{sec:dissociation}

The 50\% concordance between distributional screening and difficulty-based content assessment is the second major finding. Under S2 and S3, models concentrated 50--72\% of responses at a single position while maintaining significant difficulty-accuracy correlations ($\rho = -0.18$ to $-0.37$). The positional shortcut and content processing co-exist.

This has direct implications for validity screening. A distributional screen alone (entropy $< 0.90$) would flag S2 and S3 as invalid. This would discard responses that contain real item-level signal. The independent difficulty-accuracy criterion correctly identifies these conditions as content-engaged despite their collapsed distributions.

This parallels the distinction in clinical assessment between response-style indicators and content validity indicators. The MMPI-3's F scale and the VRIN scale capture different dimensions of invalidity. A profile can show elevated F without elevated VRIN, and vice versa. The entropy-$\rho$ dissociation in LLM evaluation reflects the same structural independence.

\subsection{Practical implications}\label{sec:practical}

Three findings have direct practical implications for red-team evaluation design.

First, increasing instruction specificity does not monotonically increase the sophistication of adversarial behaviour. The most carefully specified instruction (S5) produced the least informative responses.

Second, detection effort should focus on the intermediate-specificity regime (S2, S3). In this regime the model uses a positional shortcut while retaining enough content engagement to partially imitate genuine capability.

Third, any evaluation where a model's response distribution shows marked positional concentration should be checked against an independent content-engagement criterion before the accuracy data is interpreted. The principle is borrowed from clinical psychometrics, where it has been standard practice for decades.

\subsection{Limitations}\label{sec:limitations}

\textbf{Model scale and family.} Two 8B-parameter models from a single architectural family at Q5\_K\_M quantisation. Frontier models may resolve the S5 instruction successfully.

\textbf{Prompted setting.} All adversarial conditions are explicit instructions. This maps prompt-induced shortcuts but does not address naturalistic scheming \citep{vanderweij2025}, password-locking, or fine-tuned sandbagging \citep{meinke2024}.

\textbf{Greedy decoding.} $T=0$ only. S5 collapse may be less extreme under stochastic decoding.

\textbf{Mechanism underdetermination.} The S5 collapse is consistent with instruction-decomposition failure, representational interference, token-level priors, or decoding artefacts. Only one multi-step instruction (S5) was tested. The claim that multi-step structure drives collapse requires replication with alternative two-step phrasings.

\textbf{Difficulty as a content proxy.} The Spearman $\rho$ between item difficulty and adversarial correctness is an indirect measure of content engagement. However, the sharp contrast between S5 ($\rho \approx 0$) and S4 ($\rho \approx -0.12$ to $-0.27$) under adjacent conditions with similar accuracy levels suggests the measure captures a real difference in content processing.

\section{Conclusion}\label{sec:conclusion}

Instruction-tuned LLMs at the 8B scale resolve adversarial evaluation instructions through at least three distinct response regimes. Content-engaged compliance, shortcut with residual engagement, and extreme positional collapse. The only multi-step adversarial instruction tested coincided with the most extreme collapse, matching the model's content-absent default. This effect replicates across two models and four academic domains.

Distributional screening (response-position entropy) and content-engagement testing (difficulty-accuracy correlation) capture partially independent dimensions of response validity. A model can use a positional shortcut while retaining content sensitivity. Only the most extreme instructions produce the joint collapse that both criteria agree on. A two-stage validity screening architecture provides richer evaluation information than either criterion alone.

\section*{Acknowledgements}
Not applicable.

\section*{Data availability}
All 29,600 response records, analysis scripts, and figure-generation code are publicly available at \url{https://github.com/synthiumjp/bcb-sandbagging-pilot}. Pre-registration: \url{https://osf.io/7p64}.

\section*{Code availability}
All code is available at \url{https://github.com/synthiumjp/bcb-sandbagging-pilot}.

\section*{Declarations}

\noindent\textbf{Funding.} The author did not receive support from any organisation for the submitted work.

\noindent\textbf{Competing interests.} The author has no relevant financial or non-financial interests to disclose.

\noindent\textbf{Ethics approval.} Not applicable. No human participants were involved.

\noindent\textbf{Use of AI tools.} Claude (Anthropic) was used for analysis pipeline design, code generation, and manuscript preparation. All scientific decisions, experimental design, and interpretive claims were made by the author.

\end{document}